# Pyramid Vector Quantization for Deep Learning

Vincenzo Liguori    Ocean Logic Pty Ltd, Australia    Email: enzo@ocean-logic.com

*Abstract*—This paper explores the use of Pyramid Vector Quantization (PVQ) to reduce the computational cost for a variety of neural networks (NNs) while, at the same time, compressing the weights that describe them. This is based on the fact that the dot product between an N dimensional vector of real numbers and an N dimensional PVQ vector can be calculated with only additions and subtractions and one multiplication. This is advantageous since tensor products, commonly used in NNs, can be re-conduced to a dot product or a set of dot products. Finally, it is stressed that any NN architecture that is based on an operation that can be re-conduced to a dot product can benefit from the techniques described here.

*Index Terms*— Machine vision, Convolutional Neural Networks, Deep Learning, Vector Quantization.

## I. Introduction

In the past decade, biologically inspired artificial neural networks (NNs) have re-gained popularity thanks to the availability of ever increasing computing power required during their learning phase. Convolutional Neural Networks (CNNs)[1] are the current prevailing implementation of artificial neural networks. The efficacy of CNNs is also well known in fields from machine vision to pattern recognition and many others.

Their success in multiple fields has resulted in a wide search of methods for their efficient computation, especially at inference time, as well as methods to reduce the number of bits required to describe them. Such methods have relied on pruning or simplifying a network [2] as well as quantization of the weights (in [3] to 16 bits). Quantization has been pushed to the extreme case of binary activation functions and binary (+1,-1) weights [4][5][6]. Although the basic idea of binarized weights NNs is not new [7], in their modern incarnations they have reached a level of sophistication that has practical implication for the real world use of NNs.

Reducing the computational cost at inference time has also applications in embedded systems and low power, real time hardware implementation.

This paper will explore how PVQ, a vector quantization technique, can be used to quantize NNs weights. This simplifies the computation of NNs as well as compresses their description.

Specifically, PVQ will be applied to NNs weights and show a small degradation in performance in exchange for a more compact network. Also, since dot products with PVQ vectors can be calculated only with addition and a single multiplication, a substantial reduction of the computational cost at inference time can be achieved. In most practical cases, the single multiplication can also be eliminated.

## II. Pyramid Vector Quantization

A pyramid vector quantizer[8] (PVQ) is based on the cubic lattice points that lie on the surface of an N-dimensional pyramid. Unlike better known forms of vector quantization that require complex iterative procedures in order to find the optimal quantized vector, it has a simple encoding algorithm.

Given an integer K, any point on the surface of an N-dimensional pyramid $\hat{y}$ is such that

$$\sum_{i=0}^{N-1}|\hat{y}_i|=K \quad (1)$$

with $\hat{y}_i$ integers. The pair of integers N and K, together with (1), completely define the surface of an N-dimensional pyramid indicated here with $P(N,K)$.

In this work a particular type of PVQ will be used, known as product PVQ. Here a vector $\vec{y}\in\mathbb{R}^N$ is approximated by its norm $r=\|\vec{y}\|_2$ (also referred to as "radius" or "length" of the vector) and a direction in N-dimensional space given by the vector that passes between the origin and a point $\hat{y}$ on the surface of the N-dimensional pyramid:

$$r\frac{\hat{y}}{\|\hat{y}\|_2} \quad (2)$$

Note that the direction in N-dimensional space is effectively vector quantized. Null vectors are represented by $r=0$. The radius $r$ can also be quantized with a scalar quantizer. The vector $\hat{y}$ needs to be normalized as it does not lie on the unit hyper-sphere. Given N, increasing K increases the number of quantized directions in N-dimensional space and, hence, the quality of the approximation.

The paper[8] also includes simple algorithms to calculate the number of points $N_p(N,K)$ on the surface of the N-dimensional pyramid. It also provides algorithms to map any point on said surface to an integer $0\leq i<N_p(N,K)$ and vice-versa. Such mapping provides a much more compact representation of a surface point then a direct bit representation.

For example, for N=8 and K=4, each component $\hat{y}_i$ would naively need 4 bits (including the sign), for a total of 8x4=32 bits for the whole vector $\hat{y}$. However, because of the constraint (1), $N_p(8,4)=2816$ and, therefore, less than 12 bits are required to map any $\hat{y}\in P(8,4)$.

The mapping of $\hat{y}$ to an integer is not essential to the vector quantization of $\vec{y}$ but it can be useful in those applications where a quantized vector needs to be stored in a more compact way.

In this work, "PVQ encoding" or simply "encoding" a vector $\vec{y}$ will mean finding its closest approximation (2). "Mapping a vector to an integer" will refer to the process that associates a PVQ vector $\hat{y} \in P(N,K)$ to an integer $0 \leq i < N_p(N,K)$. Similarly for its opposite.

PVQ is suitable for quantizing Laplacian sources and, in fact, it is used to quantize transformed images (whose samples can be modeled with a source that is Laplacian or approximately so) for their compression.

The computational cost of PVQ encoding is not very high. In any case, in this paper, PVQ vectors will be considered pre-calculated constants. In other words, PVQ encoding will be understood to be performed offline and not during the inference of a NN.

## III. Dot Product

We will now look at the dot product between a PVQ vector (2) $\hat{y} \in P(N,K)$ with radius $r$ and an N-dimensional vector $\vec{x} \in \mathbb{R}^N$:

$$r \frac{\hat{y}}{\|\hat{y}\|_2} \cdot \vec{x} = \frac{r}{\|\hat{y}\|_2} \sum_{i=0}^{N-1} \hat{y}_i x_i \quad (3)$$

The author has shown in [9] that $\sum_{i=0}^{N-1} \hat{y}_i x_i$ can be calculated with exactly K-1 additions and/or subtractions and no multiplications for each possible $\hat{y} \in P(N,K)$. In this case, since we consider all PVQ vector as calculated offline, the scaling factor $\rho = \frac{r}{\|\hat{y}\|_2} \geq 0$ can also be pre-calculated and considered as a single value. In this case the dot product between a PVQ approximated vector (2) and a vector $\vec{x} \in \mathbb{R}^N$ takes K-1 addition/subtractions and one multiplication by the factor $\rho$.

## IV. Neural Networks

Detailed description of NNs is beyond the scope of this article. Suffice to say that the output y of an artificial neuron can be modeled as:

$$y = f\left(\sum_{i=0}^{N-1} w_i x_i + b\right) = f(\vec{w} \cdot \vec{x} + b) \quad (4)$$

Where $f()$ is a non-linear function (also known as activation function), $\vec{w} = (w_0, \cdots, w_i, \cdots, w_{N-1}) \in \mathbb{R}^N$ and $b \in \mathbb{R}$ are constants (known as weights and bias, respectively) and $\vec{x} = (x_0, \cdots, x_i, \cdots, x_{N-1}) \in \mathbb{R}^N$ are the inputs to the neuron. If we now concatenate $\vec{w}$ and $b$ to form $\vec{w}' = (\vec{w}, b) \in \mathbb{R}^{N+1}$ and $\vec{x}$ and 1 to form $\vec{x}' = (\vec{x}, 1) \in \mathbb{R}^{N+1}$ then (4) can be expressed as:

$$y = f(\vec{w}' \cdot \vec{x}') \quad (5)$$

Equation (5) is the one that will be used in the rest of the paper for the model of an artificial neuron.

NNs have many different architectures. For example, in fully connected networks, each neuron is connected to all available inputs (or to all previous neurons in case of multi-layer) whereas CNNs can be considered a subset of fully connected NNs. In fact, each neuron is only connected to a subset of the available input and the values of some weights are shared. In CNNs, groups of input images are mapped to groups of output images using a tensor product. The latter, once the tensor is flattened, is effectively re-conduced to a dot product. There are also other architectures, but a common theme is a dot product as in (5).

Given the ubiquity of the dot product in NNs and the advantage of performing the dot product with a PVQ vector, the following is proposed:
1. Train a NNs as usual
2. Perform PVQ on groups of its original weights
3. Test the NN with the new weights for loss of accuracy

Note that the amount of quantization can be tuned by varying the parameter K in the PVQ encoding process. A few iterations at steps 2) and 3) might be necessary to optimize the trade off between accuracy and inference performance.

This means that the weights $\vec{w}'$ in (5) are substituted by $PVQ(\vec{w}') = \rho \hat{w}'$ resulting in:

$$y = f(\rho \hat{w}' \cdot \vec{x}') \quad (6)$$

The advantage of this procedure is that every dot product on N-dimensional vectors that used to take N multiplications and N-1 additions can now be performed with K-1 additions and one multiplication. NNs with weights that are PVQ vectors will be referred to as PVQ nets.

At this point it is legitimate to ask what kind of accuracy can be expected from such approximation. We already know that PVQ performs well when the source is Laplacian or approximately so. This means in practice that PVQ should perform well for weight distributions that have a high frequency around 0 and drop off quickly for higher values (in absolute value). This is indeed what the author has observed in his experiments. It also appear to be the case for published work like [3] (see fig. 7). According to the same paper, as well as some insights of the author, $L_1$ and $L_2$ regularizations, applied to both the activities and the weights during NN training can help to sparsify the weights as well as improve the statistical properties that help PVQ encoding.

In practice, the author has observed that, for CNNs, for all convolutional layers (except the first), using $N \simeq K$ in the PVQ processing of the weights results in a drop of accuracy of a few %. For the first layer less quantization is necessary with K equal to 1.5x to 3x N. Fully connected layers seem to be more resilient to PVQ with the ratio $\frac{N}{K}$ as high as 2 to 5. It is important to realize that, with $N \simeq K$, the N multiplications are reduced to one and only N-1 additions remain.

Finally, training a NN can be formulated as an optimization problem with the weights $\vec{w}' \in \mathbb{R}^n$ as the variables to be optimized. An alternative approach would be to restrict the search space for the same optimization problem directly on the surface of the hyper-pyramid P(N,K). In other words, instead of a continuous optimization problem with variables $\vec{w}' \in \mathbb{R}^n$ followed by PVQ encoding of the weights, we would have a mixed optimization problem with $\rho \in \mathbb{R}$ and $\hat{w}' \in PVQ(N, K)$ as variables. This alternative approach is worth noting but it will not be discussed any further in this paper. A hybrid optimization technique is also possible:

1. Train a NNs as usual
2. Perform PVQ on groups of its original weights
3. Continue training as the mixed optimization problem described above

Some preliminary experiments seem to indicate that step 3) acts as a refining and improving step.

Yet another possible algorithm can be defined as K-annealing. The PVQ parameter K defines the level of quantization with larger K indicating lower quantization noise.

The mixed optimization problem is started with a high value for K. This is gradually lowered ("annealed") to the target K as the optimization proceeds.

## V. Further Optimizations

Let's consider a set of M neurons, each with its set of weights $\vec{w}_i'$ and inputs $\vec{x}_i'$:

$$y_i = f(\vec{w}_i' \cdot \vec{x}_i') \quad \text{with} \quad 0 \leq i < M \tag{7}$$

The dimensionality of each set of weights and inputs does not need to be the same (i.e. $dim(\vec{w}_i') \neq dim(\vec{w}_j')$). We already know that, if the statistics of the weights are favorable, we can approximate the set of neurons by PVQ encoding the weights and substituting them to the original. This means by $PVQ(\vec{w}_i') = \rho_i \hat{w}_i'$ resulting in:

$$y_i = f(\rho_i \hat{w}_i' \cdot \vec{x}_i') \tag{8}$$

Let $\vec{W} = (\vec{w}_0', \cdots, \vec{w}_i', \cdots, \vec{w}'_{M-1})$ and $\vec{X} = (\vec{x}_0', \cdots, \vec{x}_i', \cdots, \vec{x}'_{M-1})$ be built by concatenating the aforementioned weights and inputs. Let's also PVQ encode $\vec{W}$:

$$PVQ(\vec{W}) = \rho \hat{W} = \rho(\hat{w}_0'', \cdots, \hat{w}_i'', \cdots, \hat{w}''_{M-1}) \tag{9}$$

Note that, in general, $\hat{w}_i' \neq \hat{w}_i''$. Let's now consider the dot product:

$$\rho \hat{W} \cdot \vec{X} = \rho(\hat{w}_0'' \cdot \vec{x}_0 + \cdots + \hat{w}_i'' \cdot \vec{x}_i + \cdots + \hat{w}''_{M-1} \cdot \vec{x}_{M-1}) \tag{10}$$

Leaving aside the fact that this particular dot product has no particular meaning, we can stress once again that, as for any dot product with a PVQ vector, it can be calculated with K-1 additions or subtractions and one multiplication. This is important because it shows that the total number of additions necessary to calculate all the partial dot products $\hat{w}_i'' \cdot \vec{x}_i$ will be $\leq K-1$. More importantly (and we shall see why soon), if we PVQ all the weights $\vec{w}_i'$ concatenated together as $\vec{W}$, then the scaling factor $\rho$ will be a single scalar instead of M different ones had we PVQ encoded each $\vec{w}_i'$ separately as in (9). Now (9) can be re-written as:

$$y_i = f(\rho \hat{w}_i'' \cdot \vec{x}_i') \tag{11}$$

And, for any activation function for which

$$f(\rho x) = \rho f(x) \tag{12}$$

such as ReLU (very useful in practice), we have:

$$y_i = \rho f(\hat{w}_i'' \cdot \vec{x}_i') \tag{13}$$

In other words, if PVQ encode the weights of a group of neurons together, then, with (12) true, the scaling factor $\rho$ can "pass through" the non-linearity.

This is very important as NNs are built in stacked layers. If we PVQ encode all the weights for a particular layer together, then, with (13) true, we have:

- All the calculations for that layer, before the activation function can be done with at most K-1 addition and subtractions
- The output for all the neurons for that layer will be scaled by the same $\rho$ since $\rho$ can "pass through" the activation function as in (14)

Let's focus on the second point. For many NNs, the outputs of one layer become the inputs of the next. If all the outputs of one layer were scaled by the same factor, then the next layer will also see all its inputs scaled by the same factor. Let's see what happens when we also PVQ the weights of the next layer and assume that all the inputs are scaled by the same value, with (12) true:

$$z_i = f((\rho_1 \hat{w}_i) \cdot (\rho_0 \vec{x}_i)) = f(\rho_1 \rho_0 \hat{w}_i \cdot \vec{x}_i) = \rho_1 \rho_0 f(\hat{w}_i \cdot \vec{x}_i) \tag{14}$$

Here $\rho_1$ is the scaling factor resulting from the PVQ encoding of the current layer weights while $\rho_0$ is the one from PVQ encoding of previous layer weights (and "passed through" the activation function (12)). This can be repeated for any number of layers.

So, if we apply PVQ encoding to each layer of a NN, with a suitable activation function (12), the scaling factor $\rho$ for each PVQ vector dot product can be propagated through the network, layer by layer, up to the outputs of the network which will have a final scaling factor $\rho = \prod_{i=0}^{L-1} \rho_i$ for an L-layer NN. In other words, only the outputs of such network will have to be scaled by $\rho$. This is an important result because:

- The number of outputs in a NN is, in general, smaller and often much smaller than the number of outputs of all the neurons in the network. Therefore the impact of multiplications is limited to the outputs only.
- In many cases the output of the NN uses one hot encoding and no activation is applied. In this case the output of the NN is given by the argmax function that is not influenced by a positive scaling factor. Therefore, in this case, the latter can be completely eliminated.
- All the layers of the network can be calculated with only addition and subtractions (because they have been subject to PVQ encoding).
- For many useful NNs, the input values are integers (i.e. 8 bit pixels) and this means that all the layers can be calculated by only addition and subtraction of integer values. These are referred to as integer PVQ nets.

This is also true if the network contains Maxpool layers because, remembering that $\rho \geq 0$:

$$Max(\rho y_0, \ldots, \rho y_{N-1}) = \rho Max(y_0, \ldots, y_{N-1}) \quad (15)$$

So, the scaling factor $\rho$ will also propagate through Maxpool layers. And this also applies to convolutional as well as fully connected layers.

This means that many existing NNs, can be converted to integer PVQ nets with a substantial computational advantage.

Yet another advantage of integer PVQ nets is that, being computed only with addition and subtractions of integer values, the precision required can be easily tracked through all the layers. In fact, for deep networks, the full precision is probably not necessary and, at each layer, one can simply re-scale the values by a power of 2 (i.e. with shift operations) in order to reduce the number of bits required.

Another type of activation function that leads to great computational advantage in PVQ nets is the one for which:

$$f(\rho x) = f(x) \quad (16)$$

is true, at least when $\rho \geq 0$ which is obviously the case in the case of PVQ vector scaling factors. In this case the scaling factor $\rho$ of the PVQ encoded weights is simply "adsorbed" by the activation function and all the properties previously described also apply, without the need to "propagate" $\rho$ to the outputs as before.

An example of such activity function can be found in the binary version (with only +/-1 output values) of the $sign(x)$ function (which is ternary with -1, 0 and +1 as possible outputs):

$$bsign(x) = \begin{cases} +1 & if\ x \geq 0 \\ -1 & if\ x < 0 \end{cases} \quad (17)$$

It is possible to train multi-layer NNs with (17) as activation function and continuous weights. We can then PVQ encode the weights at each layer. For each layer, the scaling factor $\rho$ is then eliminated because of the nature of the activation function (16). The calculations are now simplified to addition and subtractions of +1 and -1 values only (except for the first layer where, if the inputs are integers, it will still be computed with integer additions and subtractions). We will refer to these nets as binary PVQ nets.

Note the difference from binarized networks such as [4] or [6]. In these all weights are either +1 or -1 whereas in binary PVQ nets a weight can be (theoretically) as large as +/-K. However, the total number of additions or subtractions in a layer is guaranteed to be $\leq K-1$. If in a particular layer of a binary PVQ net N=K, then the total number of binary additions and subtractions will be the same as in [4] but each individual weight doesn't need to be +/-1.

This is an important difference and it is worth making a simple example. Suppose we have a single neuron in a binary net with N=7 binary inputs and weights. Suppose the weights are (-1,1,1,1,-1,-1,1). The absolute value of each weight must be 1 and the dot product with the binary input will require 6 additions or subtractions. Suppose now we have a binary PVQ net with a single neuron and 7 inputs with N=K=7. A weight vector can have values different from +/-1 like (-2,1,0,0,0,2,2) or (0,0,-3,0,-2,2,0) (both respecting constraint (1)) but the dot product will still require 6 additions or subtractions.

## VI. WEIGHTS COMPRESSION

After PVQ encoding the weights can be losslessly compressed. In fact, as already mentioned, [8] describes an algorithm to map a point on a hyper-pyramid P(N,K) to an integer. Such integer can obviously be represented by a string of bits that completely and compactly describes the integer part of a PVQ vector. The scaling factor $\rho$, if required, must be quantized separately.

Unfortunately, the algorithm given in [8] is not very practical, especially when it comes to the inverse process (i.e. converting the integer number back to a PVQ vector). This is because it can involve multiple arithmetic operations on numbers thousands of bit long. However, an important advantage remains in mapping a PVQ vector to an integer, as mentioned in section II: unlike the methods that will be discussed below that result in a string of bits of unpredictable size, the method described in [8] requires $\log_2(N_p(N,K))$ bits for any $\hat{w} \in P(N,K)$.

More conventional compression techniques can still be effective and much more practical. In fact, after PVQ encoding, the values of the elements of the vector quantized vector are not all equiprobable: 0 and +/-1 values being, for example, far more likely than any others. This means that a Huffan encoding scheme is a possibility, assigning a smaller codeword to the most likely values. The drawback is the creation of a potentially very large table. This is because if we want to Huffman encode PVQ encoded weights for a NN layer, generally speaking, we are talking about thousands of dimensions. Now, with $N \simeq K$ and +/-K the largest (theoretical) value to Huffman encode, a very large table would berequired. A more practical scheme would consist in creating a Huffman table for each value whose absolute value is less than a certain value V plus an escape code. Any element of the PVQ encoded vector will then have its own Huffman code if its absolute value is less than V, otherwise the escape code is used followed by the full binary representation of all the other possible values (minus the ones covered by the Huffman code).

Golomb exponential codes are also an interesting possibility: they do not need the storage of large tables and they are suitable for encoding values whose frequency decays rapidly with their increase in magnitude (just like the values of the elements of a PVQ encoded vector).

For fully connected layers in a NN where the ratio $\frac{N}{K}$ can be as high as 5, run length encoding is a good fit as it allow less than one bit per weight for long runs of zeros. In a PVQ encoded vector with a ratio $\frac{N}{K} \simeq 5$, at least $\frac{4}{5}$ of the values are guaranteed to be zero. This is because, best case, there will be only K elements with absolute value equal to one leaving all the other elements necessarily set to zero (remember the constraint (1) that applies to all PVQ vectors).

Arithmetic encoding is also a possibility although the author is not particularly keen on this method given its difficulty to parallelize and/or randomly access a particular point in the compressed bitstream.

The similarity between the type of data obtained from image and video after transform and quantization and the PVQ encoded NN weights is an open invitation to re-use and adapt algorithms from the lossless compression stages of image and video compression standards such as JPEG, H.264 and others. These are designed to compress similar types of data where small values are more frequent.

VII. EXPERIMENTS

The author performed a series of experiments by training some simple NN using tools such as Keras[10] and Tensorflow[11] and then by PVQ encoding the original weights a whole layer at the time.

The most accurate PVQ encoding algorithm known to the author has O(NK) complexity. Since even in relative small networks it might be required to PVQ encode vectors with a dimensionality of 1,000,000 or more, it became necessary for the author to create a GPU implementation using CUDA.

Also, the author is new to the field of NN training and his results are far from the state of the art. However, the important point here is not the absolute accuracy of NN but, rather, the relatively small loss of accuracy that a NN suffers when PVQ encoding is applied to its layers. In other words, the important point here is the small drop in performance for a NN brought by PVQ encoding that is traded off for all the computational, storage and bandwidth advantages discussed.

The purpose of the experiments mentioned here is to try to ascertain, for some specific NN, what kind of PVQ encoding is possible if one is willing to tolerate an overall loss of accuracy of a few %. PVQ encoding is applied to the weights of each layer of a NN separately. Specifically, for each layer, the following steps are taken:

- Extract all the weights and biases from the given layer.
- The weights (in the form of a tensor or a matrix) are flattened and then concatenated with the biases to form a single vector of dimensionality N.
- PVQ encoding is applied to the N-dimensional vector with a given quantization parameter K, resulting a scalar $\rho$ and an integer vector $\hat{w} \in P(N,K)$.
- The vector $\rho \hat{w}$ will then be split into its weights and biases components.
- The weights vector so obtained will now be re-assembled into its original matrix/tensor shape and biases.
- The original weights and biases for the given layer will be now replaced with the new quantized ones.

Note that no further refinement or mixed optimization (as mentioned at the end of section IV) was performed to refine or improve on PVQ encoding.

| Layers | N | N/K |
|---|---|---|
| IN | 784 | - | -
| FC0 | 512 | 401,920 | 5 |
| DRP | 0.2 | - | - |
| FC1 | 512 | 262,625 | 5 |
| DRP | 0.2 | - | - |
| FC2 | 10 | 5,130 | 5 |

*Table 1: MNIST dataset NN A.*

Table 1 shows the anatomy of the NN taken from the Keras examples for the MINIST dataset [12] (indicated with A for future reference). Different layers are indicated by IN for input, FC for fully connected, CONV for convolutional, MAX for maxpool, DRP for dropout. The activation function used is ReLU.

Note that PVQ encoding with the procedure described above is only applied to layers that contain weights such as CONV and FC. For these layers, N indicates the dimensionality of the flattened vector (including biases) and the quantization parameter K is expressed as ratio with N.

For the NN shown in Table 1, the testing accuracy went from 98.27% before PVQ encoding to 95.33%. after.

| Layers | | N | N/K |
|---|---|---|---|
| IN | 3x32x32 | - | - |
| CONV0 | 3x3,32 | 896 | 1/3 |
| CONV1 | 3x3,32 | 9,248 | 1 |
| MAX | 2x2 | - | - |
| DRP | 0.25 | - | - |
| CONV2 | 3x3,64 | 18,496 | 1 |
| CONV3 | 3x3,64 | 36,928 | 1 |
| MAX | 2x2 | - | - |
| DRP | 0.25 | - | - |
| FC4 | 512 | 2,097,664 | 4 |
| DRP | 0.5 | - | - |
| FC5 | 10 | 5,130 | 1 |

*Table 2: CIFAR10 dataset NN B.*

Table 2 shows the anatomy of the NN taken from the Keras examples for the CIFFAR10 dataset [13] (indicated with B). Again, PVQ encoding is only applied to layers that contain weights. For the NN shown in Table 2, the testing accuracy went from 78.46% before PVQ encoding to 73.21%. after.

A few observations:
- After PVQ encoding, all the networks are greatly simplified and compressed, especially for fully connected layers.
- As seen also in [4][6], the first layer is the hardest to quantize. Fully connected layers seem to be the most compressible.
- Since the inputs are integers and the activation function is a ReLU, after PVQ encoding, these are essentially integer PVQ nets. Thus their inference only requires integer additions and subtractions.

More experiments were performed on these networks by changing the activation function to (17). Training of NNs is based on optimization of an objective function with variations of the gradient descent method. The latter implies the existence of the derivative of the objective function which would be essentially zero if one were to use (17) as activation function. In order to overcome this problem, Geoff Hinton in [14] suggests to use a "Straight Through Estimator" (STE) which, essentially, consists of imposing a derivative for (17) by definition:

$$\frac{d}{dx}bsign(x)=1$$

(18)

Unfortunately Keras does not support user defined activation functions and its derivatives. Therefore the author had to convert the two NNs above to Tensorflow.

After converting to Tensorflow, changing the activation function to (17) and its pseudo-derivative (18) was difficult but possible.

| Layers | N | N/K |
|---|---|---|
| IN | 784 | - | - |
| FC0 | 512 | 401,920 | 5/2 |
| FC1 | 512 | 262,625 | 5 |
| FC2 | 10 | 5,130 | 4 |

*Table 3: MNIST dataset NN C with binarized neurons.*

Table 3 shows the same NN shown in Table 1 with the ReLU activation function changed to (17) (indicated with C). The testing accuracy for this NN went from 94.14% before PVQ encoding to 91.28%. after.

| Layers | | N | N/K |
|---|---|---|---|
| IN | 3x32x32 | - | - |
| CONV0 | 3x3,32 | 896 | 2/5 |
| CONV1 | 3x3,32 | 9,248 | 1 |
| MAX | 2x2 | - | - |
| CONV2 | 3x3,64 | 18,496 | 3/2 |
| CONV3 | 3x3,64 | 36,928 | 2 |
| MAX | 2x2 | - | - |
| FC4 | 512 | 2,097,664 | 5 |
| FC5 | 10 | 5,130 | 1 |

*Table 4: CIFAR10 dataset NN D with binarized neurons.*

Table 4 shows the same NN shown in Table 2 with the ReLU activation function changed to (17). When converted to Tensorflow, even with ReLU as activation function, this NN never reached the same level of accuracy reached with Keras (with Tensorflow backend): 70.74%. With the activation function changed to (17), testing accuracy was 61.62% before and 58.54% after PVQ encoding. Some considerations:
- These are only few of the possible K assignments during PVQ encoding: others might lead to better approximation for the same level of compression.
- After PVQ encoding, these NNs are binary PVQ nets and all layers (except the first whose inputs are integers) can be calculated with additions and subtractions of binary values only.
- For layers where N/K~=1, the number of addition and subtractions of binary values is the same as in a binarized net as in [4][6]. However, for layers where N/K>1, the number of additions and subtractions of binary values is smaller than a binarized net.
- For both NNs dropout was not used as it resulted in worse results. It is possible that binarized nets are self-regularizing: training and testing accuracy were always close.

- As stressed before, the important point is the relatively small drop in accuracy compared to the gain in performance during inference.

No direct experiments on further, lossless compression of the weights have been performed. However, weights statistics after PVQ encoding have been collected and shown below. The experimental results strongly support the discussion in section VI as the weight statistics show their high compressibility.

|     | 0 | ±1 | ±2..3 | ±4..7 | Others |
|-----|---|----|-------|-------|--------|
| FC0 | 326,314 | 71,184 | 4,401 | 21 | 0 |
|     | 81.19% | 17.71% | 1.1% | 0.0052% | 0% |
| FC1 | 210,907 | 50,966 | 783 | 0 | 0 |
|     | 80.3% | 19.4% | 0.3% | 0% | 0% |
| FC2 | 4,084 | 1,032 | 14 | 0 | 0 |
|     | 79.61% | 20.12% | 0.27% | 0% | 0% |

*Table 5: PVQ weights distribution for NN A.*

Table 5 shows the distribution for the weights of NN A after PVQ encoding, layer by layer. Note that all the weights are integer values as the $\rho$ factor from PVQ encoding is not used for these NNs. Now, to get an idea of how compressible the NN PVQ weights are, let us consider a simple scheme that uses exponential Golomb codes for each value. Exponential Golomb codes will use 1 bit for 0 values, 3 bits for ±2..3, 5 bits for ±4..7, etc. So, the average for FC0 in NN A will be 0.8119+3*0.1771+5*0.011+7*0.000052=~1.4 bits/weight.

However, due to the large number of zeros, it is likely that a better result can be obtained with a simple run-length encoding scheme.

|     | 0 | ±1 | ±2..3 | ±4..7 | Others |
|-----|---|----|-------|-------|--------|
| CONV0 | 81 | 176 | 307 | 308 | 24 |
|       | 9.04% | 19.64% | 34.26% | 34.37% | 2.68% |
| CONV1 | 3,342 | 3,774 | 1,854 | 272 | 6 |
|       | 36.14% | 40.81% | 20.05% | 2.94% | 0.065% |
| CONV2 | 7,113 | 7,546 | 3,174 | 552 | 111 |
|       | 38.46% | 40.8% | 17.16% | 2.98% | 0.6% |
| CONV3 | 14,093 | 14,090 | 7,445 | 1,290 | 10 |
|       | 38.16% | 38.16% | 20.16% | 3.49% | 0.027% |
| FC4 | 1,601,481 | 468,663 | 27,167 | 23 | 0 |
|     | 76.35% | 22.36% | 1.3% | 0.0011% | 0% |
| FC5 | 1,485 | 2,426 | 1,216 | 3 | 0 |
|     | 28.95% | 47.29% | 23.7% | 0.058% | 0% |

*Table 6: PVQ weights distribution for NN B.*

Table 6 shows the distribution for the weights of NN B after PVQ encoding. Again, using a simple exponential Golomb coding scheme as outlined above on, say, CONV1 of NN B, one gets ~2.8 bits/weight. Results on the C and D NNs follow.

|     | 0 | ±1 | ±2..3 | ±4..7 | Others |
|-----|---|----|-------|-------|--------|
| FC0 | 258,662 | 127,238 | 15,976 | 84 | 0 |
|     | 64.35% | 31.66% | 3.98% | 0.021% | 0% |
| FC1 | 211,203 | 50,384 | 1,069 | 0 | 0 |
|     | 80.41% | 19.18% | 0.41% | 0% | 0% |
| FC2 | 4,123 | 988 | 19 | 0 | 0 |
|     | 80.37% | 19.26% | 0.37% | 0% | 0% |

*Table 7: PVQ weights distribution for NN C.*

|     | 0 | ±1 | ±2..3 | ±4..7 | Others |
|-----|---|----|-------|-------|--------|
| CONV0 | 140 | 241 | 299 | 173 | 43 |
|       | 15.63% | 26.9% | 33.37% | 19.3% | 4.8% |
| CONV1 | 3,168 | 3,835 | 2,058 | 185 | 2 |
|       | 34.26% | 41.48% | 22.25% | 2% | 0.022% |
| CONV2 | 8,602 | 7,783 | 2,079 | 32 | 0 |
|       | 46.51% | 42.08% | 11.24% | 0.17% | 0% |
| CONV3 | 20,649 | 14,202 | 2,076 | 1 | 0 |
|       | 55.92% | 38.46% | 5.62% | 0.0027% | 0% |
| FC4 | 1,686,451 | 402,980 | 8,233 | 0 | 0 |
|     | 80.4% | 19.21% | 0.39% | 0% | 0% |
| FC5 | 1,787 | 2,182 | 1,033 | 120 | 8 |
|     | 34.83% | 42.53% | 20.14% | 2.34% | 0.156% |

*Table 8: PVQ weights distribution for NN D.*

### VIII. Hardware Considerations

PVQ encoding of NN weights brings many advantages to their hardware implementation such as reducing computational as well as bandwidth and storage cost. This is especially true for binary PVQ nets. Table 5 in [6] shows the advantage in hardware implementation in reducing operations from floating point to integer. The underlining assumption in this section is that PVQ encoding of weights happens offline. Thus, the number and position of zero coefficients in a PVQ vector are known in advance and they can be excluded from any calculation.

Fig.1 shows a couple of possible serial architectures for the dot product in PVQ nets. There's also an INIT signal to clear the Acc content for the first product.

The one on the left uses a multiplier to multiply a PVQ weight $\hat{w}_i$ with an input $x_i$. The result is then accumulated in Acc. This is the classical way of calculating a dot product in hardware and, with the assumption made above of $\hat{w}_i \neq 0$, it will take K clock cycles at most.

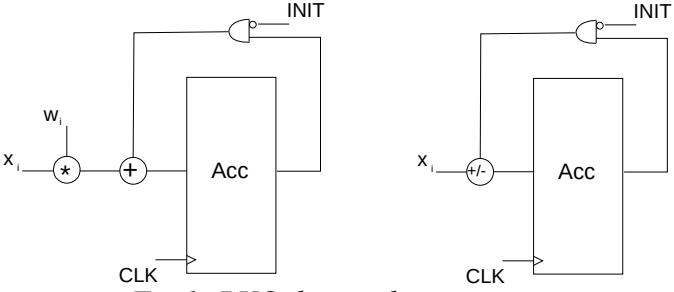

*Fig 1: PVQ dot product circuits.*

The architecture on the right, exploits the properties of a PVQ vector and it works by adding $x_i$ to Acc $\hat{w}_i$ times if $\hat{w}_i > 0$ and subtracting it $|\hat{w}_i|$ times if $\hat{w}_i < 0$ (the control signal for the add/sub is not shown). It will take exactly K clock cycles, regardless of the PVQ encoded weights. This architecture would appear to be the best because of the lack of the multiplier. This is generally the case. However, we know from the experiments that, even with $N \simeq K$, up to 1/3 for the PVQ weights is zero. This allows to calculate the dot product in less cycles in the architecture with the multiplier whereas the other one will always take K cycles. This can be an advantage in some cases, especially if we consider that the PVQ weights $\hat{w}_i$ are small in magnitude resulting in a simple multiplier (see weight statistics in the previous section). The advantage is even more pronounced in the case of fully connected layers with even more $\hat{w}_i = 0$. Finally we notice that a useful activation function as ReLU can be easily implemented at the output of either architecture in Fig.1. In fact, assuming that Acc represents numbers in two's complement, all that is needed is the sign bit of Acc to force the output to zero if negative. This can easily be done with AND gates controlled from the sign bit.

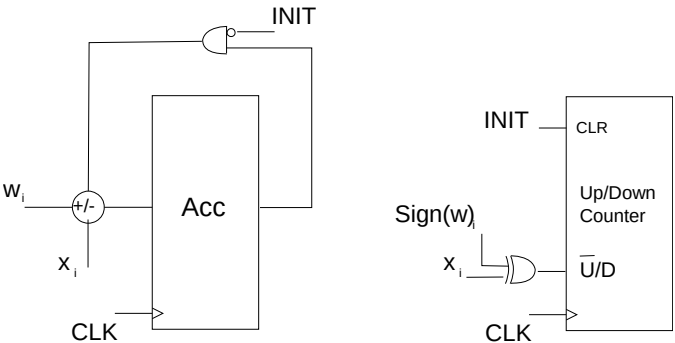

*Fig 2: Binary PVQ dot product circuits.*

We can now examine two architectures in Fig. 2 that are suitable for binary PVQ nets. They assume that their input $x_i$ is binary with 0 indicating $x_i = 1$ and 1 for $x_i = -1$.

The architecture on the left will accumulate PVQ weights with the sign inverted if $x_i = -1$ (the add/sub is controlled by $x_i$). This will take K cycles at most.

The architecture on the right is an up/down counter that can be set to zero by INIT signal. The counter will increment when U/D input is zero, decrement otherwise. If the PVQ weight $\hat{w}_i > 0$ then 0 will be input $\hat{w}_i$ times, otherwise 1 will be input $|\hat{w}_i|$ times. The XOR gate will perform a sign product with $x_i$: if the sign is the same the counter will be incremented, decremented otherwise. In other words, the counter will add or subtract $\hat{w}_i$ to its value, depending on the sign of $x_i$. This will take exactly K cycles.

Similarly to the other architectures, there is a trade off between complexity and speed.

We also note that the activation function (17) needs no circuit for its implementation: it's simply the sign bit of the Acc/counters.

Also, in FPGA implementations, we can exploit the LUT architecture to pack multiple sums of products as shown in Fig.3.

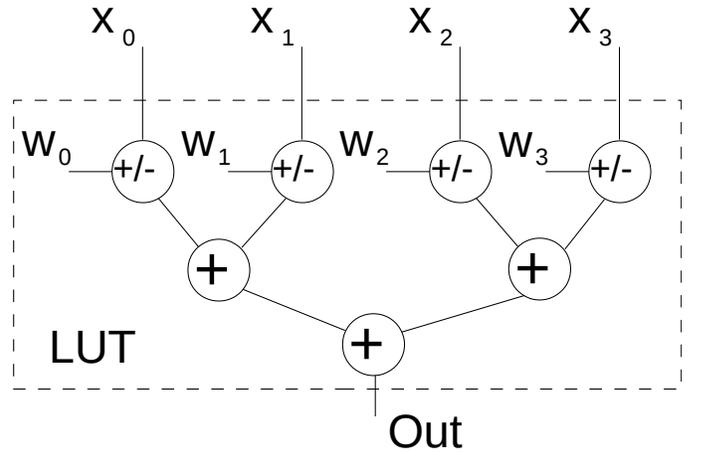

*Fig 3: Dot products for binary PVQ nets as LUTs.*

Due to the binary nature of the inputs in binary PVQ nets, multiple sums of product can be pre-calculated in LUTs. Modern FPGA have 6 input LUTs that are user programmable (even dynamically, during the computation) and can pack 6 partial sums of product as a bitslice. The number of LUTs will depend on the required precision of the output. Other FPGA families have abundant small memories that can be used for a similar purpose with an even larger number of binary inputs. The partial sums still need to be added up to form the final dot product but a lot of the computation is taken care by the LUT.

All the architectural elements described can be clearly parallelized. Some parallel architectures, as suggested in section IV in [9] can be more practical for binary PVQ nets. For example, the crossbar there described is clearly simpler for binary value and it can be implemented with Clos networks [15] either in full or in a time-multiplexing fashion.

Finally, for binary PVQ nets, the Maxpool non-linearity is simply implemented with:

$$Max(x_0, \ldots, x_{N-1}) = x_0 \text{ AND } x_1 \ldots \text{ AND } x_{N-1} \quad (20)$$

where $x_0, \ldots, x_{N-1}$ are binary values with the convention described above to represent +/-1 values.

## IX. Conclusion and Future Work

This paper has shown how to use PVQ to vector quantize NN weights and then use the properties of PVQ encoded vectors to simplify the NN inference. In particular:
- PVQ encoding the weights substantially reduces the number of bits required per weight.
- If the NN is built with activation functions like ReLU and non linearity such as Maxpool, then it can be inferred with addition and subtraction only.
- If the NN is built with activation functions like bsign(x), then it can be mostly inferred with addition and subtractions of binary values.
- This reduction in computational cost and storage is very important for low power implementations in hardware for embedded system.

Future work will focus on what can be done during training to improve results after PVQ encoding as well as post-quantization optimization steps. Hardware implementations of PVQ nets will also be created.